# CHATBOTS AS FLUENT POLYGLOTS:
# REVISITING BREAKTHROUGH CODE SNIPPETS

David Noever[1] and Kevin Williams[2]
PeopleTec, 4901-D Corporate Drive, Huntsville, AL, USA, 35805
[1] david.noever@peopletec.com  [2]kevin.williams@peopletec.com

## ABSTRACT

*The research applies AI-driven code assistants to analyze a selection of influential computer code that has shaped modern technology, including email, internet browsing, robotics, and malicious software. The original contribution of this study was to examine half of the most significant code advances in the last 50 years and, in some cases, to provide notable improvements in clarity or performance. The AI-driven code assistant could provide insights into obfuscated code or software lacking explanatory commentary in all cases examined. We generated additional sample problems based on bug corrections and code optimizations requiring much deeper reasoning than a traditional Google search might provide. Future work focuses on adding automated documentation and code commentary and translating select large code bases into more modern versions with multiple new application programming interfaces (APIs) and chained multi-tasks. The AI-driven code assistant offers a valuable tool for software engineering, particularly in its ability to provide human-level expertise and assist in refactoring legacy code or simplifying the explanation or functionality of high-value repositories.*

## 1. INTRODUCTION

The latest generation of artificial intelligence (AI) and chat applications [1-13] shows particular promise as software generators [4,11], presenting a new interactive way to learn complex coding principles [6], comment on existing code in multiple languages [8], and generally serve as coding assistants [8-12]. Recent efforts by OpenAI have put large language models (LLMs) into public access [1-2]. As an experimental platform, particularly for understanding software principles, its interactive chat [1] simulates a vast knowledge base, expert role-playing, and long-term memory spanning 8000 tokens, or approximately 20-25 pages of generated text. Several tests or benchmarks, such as QuixBugs [8] and HackerRank [12], have demonstrated the potential of generative coders as software assistants [10]. A recent review from the University of Washington and Microsoft Research [14] estimated that 1.2 million coders currently use OpenAI's copilot for tasks formerly requiring searches, such as code completion, commentary, or bug detection.

The present research seeks to understand the chatbot interface as a way for programmers to ask diverse and challenging questions to a knowledge base trained on internet-scale data but contextualized across both domain and style expertise that separates its writing quality from previous personal assistants. One observation to highlight in this analysis stems from the obviousness of history in retrospect. While a web user today might see the rise of computer codes as an inevitable consequence of networked nodes communicating securely, the inventors of the last fifty years have no such foresight into the future of image-based browsers, secure email, or eCommerce sites, or malware hurdles. In this context, applying an LLM to revisit historically significant code innovations provides a novel benchmark for future improvements in coding assistants.

## 2. METHODS

The experimental approach builds on the survey of software history's most consequential computer code [15]. As shown in Figure 1, one of the latest contributions to influential code is the transformer architecture (first proposed by Google Brain, Google Research, and the University of Toronto) [16], subsequently adopted by the AI community to build ground-breaking natural language [13] and vision models. In the example, OpenAI's latest chat interface [1] answers a question to code itself by creating a large language model unit that can scale to the billions of parameters used to build ChatGPT.

The paper explores this hypothesis by surveying other historical code snippets that prove both small enough to prompt the chat interface as viable questions and sufficiently consequential to require elaboration. The 2019 collaboration

```
write a single python function to train a large language model using repetitive uses of
transformer blocks to build a text generator

Here is a sample function that trains a large language model using repetitive transformer
blocks to build a text generator:

                                                                        Copy code

def train_transformer_text_generator(data, model, optimizer, num_blocks,
num_epochs):
    for epoch in range(num_epochs):
        for block in range(num_blocks):
            for input_text, target_text in data:
                # Run the input text through the model
                output = model(input_text)
                # Calculate the loss between the output and target text
                loss = loss_function(output, target_text)
                # Backpropagate the loss and update the model weights
                loss.backward()
                optimizer.step()
            print(f'Epoch {epoch+1}/{num_epochs} complete')
    print('Training complete')
```

Figure 1. Example coding task to self-invent a large language model.

between Arizona State University, Slate, and New America highlighted short lines of code or concise snippets in their article "*The Lines of Code That Changed Everything"* [15]. Based on 75 respondents, "pick the pieces of code that had a huge influence or warped our lives." The results showcased 36 example snippets from the first code (1725) to the Boeing 737 Max takeoff and landing software (2017) that triggered nosedives and killed hundreds.

Tasks 1-19 defines code snippets in 19 sample problems surveyed by Future Tense [15] as small code sections that changed the modern technological world over a half-century from 1961 to 2014. The experimental format includes defining the task, presenting ChatGPT with a code snippet, and then directing OpenAI's model to elaborate or explain this breakthrough software step. Appendices A-B provide examples of how detailed a coding chatbot can go for analyzing software snippets when repeatedly questioned or probed for additional suggestions. In some cases, the code is famous and recognizable enough that a Google search might similarly point the user to a human-curated interpretation of the code function on another assistance site like StackOverflow.com or StackExchange. In other cases, the highly obfuscated code of some encryption algorithms receives deep analysis that even expert coders in multiple languages might struggle to define. For example, the calculation originally proposed in Bitcoin's probability of compromise or the export-controlled code that spawned the launch of secure e-commerce.

Our approach treats ChatGPT as a new scientific instrument that might turn out historically to rival the telescope or microscope in its future incarnations when AI provides increasingly sophisticated access to not just software solutions but general capabilities for problem-solving or scientific and mathematical proofs [17]. We apply this new instrument to 19 challenge problems with a minimal hint to ChatGPT regarding their ultimate historical significance. Some code snippets correspond to image compression (JPEG), which enables the online transition from text to image-based browsing. Other code sections only hint at their ultimate role in computing museums as the first malicious worm (Morris), one-line virus (fork bomb), or pernicious security vulnerability (Heartbleed) that, upon first discovery, rendered two-thirds of the websites vulnerable overnight.

The format of the paper first summarizes the task as a benchmark problem to solve. The prompt to ChatGPT asks minimally, "what does this code do?" or "comment on this obfuscated function." We refer to any hints using the term glossator, meaning to "speak, talk or chat" and derived from Greek as even "speaking in tongues" [18]. The heuristic term underscores that, at present, these models are so large and broad that few can deeply understand whether the model technically knows what it is predicting as the next token in a long-trained memory chain or whether it genuinely can be probed for increasingly detailed analysis [19] that rivals the best computer experts in multiple languages. One notable and accepted skill for ChatGPT is its robust capabilities as a software polyglot. We demonstrate that a relatively obscure sampling of 8,945 computing languages [20] can translate ground-breaking code to underserved examples in Curry, Fish, and Glee [21]. For some languages like COBOL or FORTRAN, this translation ease may provide early profitable applications to modernize legacy code that still dominates the financial industry (COBOL) [22] or high-performance scientific libraries (FORTRAN) [23].

## 3. RESULTS

The work presents each task as an initial challenge of known importance and then revisited with coding assistance from ChatGPT. In all cases, the chat interface provides sufficient context [24] that if the initial response proves less insightful, further probing reveals that the LLM effectively comments, explains, or implements as a software polyglot.

### 3.1 CONSEQUENTIAL CODE EXPLORATION

Task 1 asks ChatGPT to explain the Assembly code underpinning the first distributed video game, MIT's version of Spacewars!

**Task 1:** *Revisit the first distributed video game in Assembly (1961), MIT DEC PDP-1 Computer, as demonstrated regularly at the Computer History Museum*

| **Prompt and code:** What does this code do?<br><br>hp1,  dap hp2<br>  count i ma1, hp2<br>  law hp3    / next step<br>  dac i ml1<br>  law 7<br>….<br>(etc.) | **Glossator:** It plays spacewars in which two players control spaceships, the needle and the wedge, and engage in a one-on-one space battle while avoiding a gravity well of a star at the center of the screen. |
|---|---|
| | **Response:** Based on this additional context, it looks like the code you provided is handling the movement of one of the spaceships in the game, possibly the "needle" spaceship. It appears to be using some randomness in its movement, and there are references to certain conditions (such as "ml1" and "mb1") that may be relevant to the game's mechanics. It is also possible that the code is interacting with other parts of the game, such as handling the display of the spaceship or processing input from the player controlling the spaceship. Without more information about the specific game mechanics and the purpose of the different variables and instructions in the code, it is difficult to say exactly what the code is doing. |
| | **Glossator:** can you translate it to a python spacewars game doing approximately the same task |
| | **Response:** Sure, here is a rough translation of the code you provided into Python that accomplishes roughly the same task:   *(Appendix A code in python)* |

Task 2 asks ChatGPT to explain the COBOL code underpinning the early origins of a networked email system.

**Task 2:** *Revisit the origins of email (1965), a formal command system for sending, receiving, and displaying small digital messages*

| Prompt and code: Explain what this code does<br>WHENEVER<br>A(1).E.FENCE.OR.A(2).E.FENCE.OR.A(3).E.FENCE<br>PRFULL.($'R'1INSTRUCTIONS:$)<br>PRFULL.($ '4MAIL NAME1 NAME2 PROB1<br>PROG1 PROB2 PROG2 ...$)<br>PRFULL.($ WHERE '=NAME1 NAME2'= IS THE<br>FILE TO BE MAILED,$)<br>PRFULL.($ AND '=PROBN PROGN'= ARE<br>DIRECTORIES TO WHICH '8$,<br>1  $IT IS TO BE SENT.'B$)<br>CHNCOM.(0)<br>END OF CONDITIONAL | **Glossator:** its an electronic mail system |
|---|---|
| | **Response:** It looks like this is a code written in a variant of the COBOL programming language. The code appears to be defining a conditional statement that will be executed whenever the condition "A(1).E.FENCE.OR.A(2).E.FENCE.OR.A(3).E.FENCE" is true. The condition appears to be checking if certain values ("A(1)", "A(2)", "A(3)") are equal to the value "FENCE". If the condition is true, the code will execute the following three statements: *(continued, Appendix B example)* |
| | **Glossator:** translate the code to python to accomplish a similar thing |
| | **Response:** Here is a rough translation of the code you provided into Python that accomplishes roughly the same thing: *(Appendix B code in python)* |

Task 3 asks ChatGPT to explain the Assembly code underpinning the memory manager and task buffer from the Apollo 11 computer and lunar lander.

**Task 3:** *Revisit the Apollo 11 Lunar Module's BAILOUT Code (1969), the code that kept the lunar module's computer from running out of space in space*

| Prompt and code: can you tell me what this code does?<br>```<br>POODOO    INHINT<br>     CA  Q<br>     TS  ALMCADR<br><br>     TC  BANKCALL<br>     CADR VAC5STOR  # STORE ERASABLES<br>FOR DEBUGGING PURPOSES.<br><br>     INDEX ALMCADR<br>     CAF  0<br>ABORT2   TC  BORTENT<br><br>OCT77770 OCT  77770    # DONT MOVE<br>     CA  V37FLBIT  # IS AVERAGE G ON<br>     MASK  FLAGWRD7<br>     CCS A<br>  TC  WHIMPER -1  # YES. DONT DO<br>POODOO. DO BAILOUT.<br><br>     TC  DOWNFLAG<br>     ADRES  STATEFLG<br><br>     TC  DOWNFLAG<br>     ADRES  REINTFLG<br><br>     TC  DOWNFLAG<br>     ADRES  NODOFLAG<br><br>     TC  BANKCALL<br>``` | **Glossator:** Its part of the memory manager for the Apollo lunar lander guidance system |
|---|---|
| | **Response:** This code appears to be written in a variant of the Assembly programming language and appears to be for use in a guidance or navigation system of some kind. It seems to be setting a flag to indicate that a certain operation (called "POODOO") has been initiated, and it is also storing certain data for debugging purposes. It appears to be checking if a certain condition (indicated by the "V37FLBIT" variable) is met, and if it is not met, the code sets some flags to specific values and calls a subroutine called "MR.KLEAN" before exiting. Without more context, it is difficult to say exactly what the code is doing or what the variables and subroutines in the code represent. |
| | **Glossator:** when the computer runs out of space or overflows, the program triggers BAILOUT to prioritize more important data |
| | **Response:** It sounds like the "BAILOUT" operation is a mechanism that is used to free up memory space or to prevent buffer overflows in the Apollo lunar lander guidance system. It may be triggered when the computer runs out of space or when the memory usage of the guidance system exceeds a certain threshold. When "BAILOUT" is triggered, the program may prioritize certain data or operations as more important and temporarily halt or suspend less important tasks or data in order to make room in the memory. The specific behavior of the "BAILOUT" operation and the |
| ```<br>        CADR    MR.KLEAN<br>        TC      WHIMPER<br>``` | criteria used to determine which tasks or data are prioritized may depend on the specific implementation of the memory manager. |

Task 4 asks ChatGPT to explain the familiar code underpinning the programming introduction, hello world, in multiple obscure languages, including Glee, Fish, and Curry.

**Task 4:** *Revisit Hello World Code (1972), the phrase that has introduced generations to code*

| **Prompt and code:** explain what this code might be used for? `main( ) { printf("hello, world\n"); }` | **Glossator:** rewrite it in Glee |
|---|---|
| **Response:** This code is a simple example of a C program that prints the text "hello, world" to the console. When the program is compiled and run, it will output the text "hello, world" followed by a newline character. This code is often used as a starting point for learning how to program in C, or as a simple example to test that a C compiler is working correctly. | **Response:** Here is the equivalent code in the Glee programming language:<br><br>```<br>main() {<br>    print("hello, world\n");<br>}<br>```<br>This code defines a function called "main" that takes no arguments and outputs the text "hello, world" to the console, followed by a newline character. In Glee, functions are defined using the "{" and "}" characters, and the "print" function is used to output a string to the console. The "\n" character is used to represent a newline character in a string. |
| **Glossator:** rewrite the code in the Curry language | **Glossator:** rewrite it in Fish |
| **Response:** Here is the equivalent code in the Curry programming language:<br><br>`main = putStrLn "hello, world"`<br><br>This code defines a function called "main" that takes no arguments and outputs the text "hello, world" to the console, followed by a newline character. In Curry, functions are defined using the "=" operator, and the "putStrLn" function is used to output a string to the console followed by a newline character. | **Response:** Here is the equivalent code in the Fish programming language:<br><br>```<br>function main<br>    echo "hello, world"<br>end<br>```<br>This code defines a function called "main" that outputs the text "hello, world" to the console. In Fish, functions are defined using the "function" keyword, and the "echo" command is used to output a string to the console. The "end" keyword is used to indicate the end of the function definition. |

Task 5 asks ChatGPT to explain the buffer overflow attack and various potential fixes for memory managers.

**Task 5:** *Revisit the Null Terminated String (1972), the most catastrophic design bug in the history of computing*

| **Prompt and code:** tell me what might go wrong with the following C code:<br>  `char yellow[26] = {'y', 'e', 'l', 'l', 'o', 'w', '\0'};` |
|---|
| **Response:** Here are a few things that might go wrong with the given C code:<br><br>If the array "yellow" is accessed outside of its bounds, it could lead to a buffer overflow or segmentation fault.<br>If the array "yellow" is not null-terminated, it could cause issues with functions that expect null-terminated strings. |
| If the array "yellow" is accessed after it goes out of scope (e.g. after the function it is defined in returns), it could lead to undefined behavior.<br>If the array "yellow" is not properly initialized, it could contain garbage values that might cause issues if the array is used for a specific purpose.<br>There may be other issues that could arise with this code as well, depending on how it is used and the specific environment it is running in. |
| **Glossator:** how could it be exploited in a buffer overflow attack |
| **Response:** In the case of the code you provided, a buffer overflow attack could potentially occur if the array "yellow" is used to store a string that is longer than 7 characters, and that string is passed to a function that does not properly check the length of the string before copying it into a fixed-size buffer. If the string is longer than the buffer, the extra data will overwrite adjacent memory locations, potentially allowing an attacker to execute malicious code or access sensitive data. |
| **Glossator:** in single line of code, can i fix it? |
| **Response:** Here is a single line of code that you could use to fix the issue of the array being too small to hold the null terminator:<br><br>```char yellow[27] = {'y', 'e', 'l', 'l', 'o', 'w', '\0'};```<br><br>This code declares an array "yellow" with 27 elements, which is large enough to hold the 6 characters in the string "yellow" plus the null terminator. This will prevent the array from overflowing and help to avoid buffer overflow vulnerabilities.<br><br>However, keep in mind that this is just a simple fix for the specific issue of the array being too small to hold the null terminator. There are other potential issues with the code that may still need to be addressed, such as the possibility of the array being used to store strings that are too long or the array being accessed using out-of-bounds indices. To fully address all potential issues with the code, you may need to use additional lines of code or implement more robust input validation and error handling. |

Task 6 asks ChatGPT to explain the rounding up and down errors that biases many addition and multiplication steps.

**Task 6:** *Revisit the Vancouver Stock Exchange's Rounding Error (1982), the minor decimal distinction with a major cost causing the electronic stock index to lose half its original value after 23 months in operation*

| |
|---|
| **Prompt and code:** How could these two lines of code produce drastically different output "return floor(value)" vs. "return round(value)" when doing many addition operations |
| **Response:** The "floor" function rounds a number down to the nearest integer, while the "round" function rounds a number to the nearest integer using standard rounding rules. These functions can produce different results when applied to the same number |
| **Glossator:** explain how it could affect the financial markets if implemented to evaluate the electronic index value summed over many trades |
| **Response:** For example, if the "floor" function is used to round the index value down to the nearest integer, it could result in the reported index value being slightly lower than it would be if the "round" function or another rounding method was used. This could potentially lead to discrepancies in the reported prices of the underlying assets and could affect the decisions of traders and investors. Overall, it is important to carefully consider the implications of using any rounding function in the financial markets, and to choose a method that is transparent, accurate, and fair to all market participants. |

Task 7 asks ChatGPT to explain the first significant cyberattack based on the Morris Worm.

**Task 7:** *Revisit the Morris Worm (1988), what has been called the "first major attack on the Internet." Approximately 10 percent of the 60,000 computers connected to the internet were hit, causing millions of dollars of damage and leading the New York Times to print the words "the Internet" for the first time.*

| Prompt and code: Please explain what the following code does:<br>```<br>checkother()           /* 0x57d0 */<br>{<br>    int s, l8, l12, l16, optval;<br>    struct sockaddr_in sin;    /* 16 bytes */<br><br>    optval = 1;<br>    if ((random() % 7) == 3)<br>      return;              /* 612 */<br><br>    s = socket(AF_INET, SOCK_STREAM, 0);<br>    if (s < 0)<br>      return;<br>``` | **Response:** This code appears to be a function in the C programming language that creates a socket and sets a socket option. A socket is a communication endpoint that is used to send and receive data over a network. |
|---|---|
| | The socket is created using the "AF_INET" domain (which stands for "Internet Address Family"), the "SOCK_STREAM" type (which indicates that the socket will use a stream-based protocol, such as TCP), and a protocol value of 0 (which specifies that the default protocol should be used). If the "socket" function is successful, it returns a socket descriptor that is stored in the "s" variable. If the "socket" function fails, it returns a negative value and the function immediately returns. |

Task 8 asks ChatGPT to explain the image compression algorithm underpinning the JPEG image format.

**Task 8:** *Revisit the JPEG format (1992), what the Joint Photographic Experts Group published specifications for a standard—the JPEG—to make image files smaller, using royalty-free lossy compression, and enable internet transport without raw or bulky data transfer*

| Prompt and code: Please comment on each line of the following: *(JPEG code without commentary)*.<br>**Response:** This code appears to implement a function that performs a discrete cosine transform (DCT) on an input array of doubles. The DCT is a mathematical operation that is used to transform a signal from the time domain to the frequency domain.<br><br>```<br>double *NaiveDct_transform(double vector[], size_t len) {<br>  // This function takes an array of doubles and a size_t as input and returns a pointer to an array of doubles.<br><br>  if (SIZE_MAX / sizeof(double) < len)<br>    // If the length of the input array is larger than the maximum size that can be stored in a size_t,<br>    // return NULL.<br>    return NULL;<br><br>  double *result = malloc(len * sizeof(double));<br>  // Allocate memory for the result array using malloc. The size of the array is len * sizeof(double).<br><br>  if (result == NULL)<br>    // If the memory allocation fails, return NULL.<br>    return NULL;<br><br>  double factor = M_PI / len;<br>  // Calculate the value of the factor that will be used in the loop below.<br>``` | **Response:** *(continued)*<br><br>```<br>  for (size_t i = 0; i < len; i++) {<br>    // Loop through the elements of the result array.<br><br>    double sum = 0;<br>    // Initialize the sum to 0.<br><br>    for (size_t j = 0; j < len; j++) {<br>      // Loop through the elements of the input array.<br><br>      sum += vector[j] * cos((j + 0.5) * i * factor);<br>      // Add the product of the current element of the input array and the cosine of the angle<br>      // (j + 0.5) * i * factor to the sum.<br>    }<br><br>    result[i] = sum;<br>    // Set the current element of the result array to the sum.<br>  }<br>  return result;<br>  // Return the pointer to the result array.<br>}<br>``` |
|---|---|

Task 9 asks ChatGPT to explain the basic image algorithm underpinning the first web browser, Mosaic.

**Task 9:** *Revisit the Mosaic browser (1993), what enabled non-textual birth of the web*

| Prompt and code: Please explain what the following code might be used for | **Response:** This code appears to define a function called "MakeImage" that takes several arguments as input: a pointer to a "Display" structure, a pointer to |
|---|---|

| ```
MakeImage(dsp, data, width, height, depth,
img_info, clip)
  Display *dsp;
  unsigned char *data;
  int width, height;
  int depth;
  ImageInfo *img_info;
  int clip;
{
  int linepad, shiftnum;
  int shiftstart, shiftstop, shiftinc;
  int bytesperline,bpp;
  int temp;
  int w, h;
  XImage *newimage;
  unsigned char *bit_data, *bitp, *datap;
  Visual *theVisual;
  int bmap_order;
  unsigned long c;
  int rshift, gshift, bshift;

#ifdef NEW
  switch(bpp=bits_per_pixel(dsp,depth))
#else
  switch(depth)
``` | an array of unsigned char (which is likely to be image data), integers representing the width and height of the image, an integer representing the depth of the image, a pointer to an "ImageInfo" structure, and an integer "clip." The function appears to create an image from the input data and display it on a screen. The rest of the function appears to use the input data and the variables declared earlier to create an "XImage" structure, which is a data structure used to represent images in the X Window System. The image is then displayed on the screen using one or more X Window System functions.<br>**Glossator:** could one use this code to build a web browser<br>**Response:** In order to build a web browser, you would need to include additional code and libraries that can handle tasks such as downloading and parsing web content, rendering HTML and CSS, and handling user input and interactions. You would also need to design a user interface and integrate it with the core browser functionality.<br><br>Building a web browser is a complex task that requires a deep understanding of many different technologies and programming languages. It would not be practical to attempt to build a browser using only the code you provided. |
|---|---|

Task 10 asks ChatGPT to explain the invisible pixel tracker that enabled social media, advertising models.

**Task 10:** *Revisit the tracking pixel (1993), what enabled web usage tracking with an invisible pixel*

| **Prompt and code:** what does this html code do? `` | **Response:** This HTML code defines an image element that is used to track page views on a website. The image has a height and width of 1 pixel and is set to be hidden using the "style" attribute. The "src" attribute specifies the URL of a tracking script hosted by Facebook. When a user visits a webpage that includes this image element, their web browser will request the image from the specified URL and send information about the page view to Facebook. Facebook can then use this information to track the number of page views on the website and to gather other data about the users visiting the site. Overall, this code is used to track page views on a website and collect data about the users visiting the site. It is often used by website owners and advertisers to gather information about their users and to improve the targeting of ads and other content.<br>**Glossator:** How could a user block this code from tracking them<br>**Response:** Use a web browser with built-in tracking protection |
|---|---|

Task 11 asks ChatGPT to explain the link sharing code for early Wikipedia compilations of large crowd-sourced knowledge bases.

**Task 11:** *Revisit the Wiki (1994), what paved the way for Wikipedia*

| Prompt and code: what does this code do?<br><br>```perl
sub AsLink {
   local($num) = (@_);
   local($ref) = $old{"r$num"};
   defined $ref
       ? ($ref =~ /\.(?:gif|jpg|jpeg|png)$/i
           ? ""
           : "<a href=\"$ref\">[$num]<\/a>")
       : "[$num]";
}
``` | **Response:** This code appears to define a function in the Perl programming language called "AsLink" that takes a single argument, called "$num". The function appears to be used to create a link or an image tag for a given reference. |
|---|---|

Task 12 asks ChatGPT to recognize the javascript one-liner that opens a popup browser window.

**Task 12:** *Revisit the popup ad (1995), what adware brought as the internet annoyance*

| **Prompt and code:** what does this code do?<br>`window.open('https://www.example.com/')` | **Response:** This code is written in JavaScript and appears to open a new window or tab in a web browser and navigate to the URL. This code could be used to create a link or button that, when clicked, opens a new window or tab and navigates to a specified website. It could also be used in other contexts, such as in a script that automatically opens a new window or tab when a webpage loads. |
|---|---|

Task 13 asks ChatGPT to explain the RSA encryption code that enabled e-commerce sites to take early payments in its obfuscated format.

**Task 13:** *Revisit the secure e-commerce enabling RSA code (1995), how the Rivest–Shamir–Adleman algorithm became a secret that could not be exported or shown to a foreign national*

| **Prompt and code:** Please comment the following perl code one-liner and explain what it does:<br><br>#!/bin/perl -s-- -export-a-crypto-system-sig -RSA-3-lines-PERL<br>$m=unpack(H.$w,$m."\0"x$w),$_=`echo "16do$w 2+4Oi0$d*-^1[d2%Sa 2/d0<X+d*La1=z\U$n%0]SX$k"[$m*]\EszlXx++p\|dc`,s/^.\|\W//g,print pack('H*',$_)while read(STDIN,$m,($w=2*$d-1+length($n)&~1)/2) | **Response:** This is a one-line Perl script that appears to implement some kind of cryptographic system. The script uses several advanced features of the Perl language and may be difficult to fully understand without some knowledge of Perl and cryptography. Based on the use of cryptographic functions and the structure of the script, it is likely that the script is intended to perform some kind of encryption or decryption on its input data. |
|---|---|

Task 14 asks ChatGPT to explain the Google PageRank algorithm that organized link priorities and relevance to user queries.

**Task 14:** *Revisit the Google's PageRank Algorithm (1996), how the most relevant links could be sorted based on the ratio of higher incoming links*

| **Prompt and code:** what does this code do?<br>`import numpy as np` | **Response:** This code defines a function called pagerank that computes the |
|---|---|

```
def pagerank(M, num_iterations=100, d=0.85):
    N = M.shape[1]
    v = np.random.rand(N, 1)
    v = v / np.linalg.norm(v, 1)
    iteration = 0
    while iteration < num_iterations:
        iteration += 1
        v = d * np.matmul(M, v) + (1 - d) / N
    return v
```

PageRank of a given matrix M. The PageRank of a matrix is a measure of the importance of each element in the matrix, with higher values indicating more importance. The PageRank is often used to rank the importance of websites in search engines.

Task 15 asks ChatGPT to explain the one-line virus that acts the first distributed fork bomb in its obfuscated format.

**Task 15:** *Revisit the one line virus (2002), how a fork bomb works*

| **Prompt and code:** what does this code do? `:(){ :\|:& };:` | **Response:** It looks like this code is actually a series of Linux shell commands separated by semicolons. It is not a valid Python code.<br><br>The first command, :, is a shell builtin that does nothing. The second command, ( ), creates a subshell and runs the commands inside it. The third command, { :\|: & }, runs the command : in the background and pipes its output to the command : using the \| operator. The fourth command, ;, is a command separator that allows multiple commands to be run on the same line. The fifth command, :, is a shell builtin that does nothing.<br><br>Overall, this code does not perform any meaningful actions and would have no effect if it were executed in a shell.<br>**Glossator:** could it be a fork bomb?<br>**Response:** It is possible that this code could be used to create a fork bomb, which is a type of denial of service (DoS) attack that can cause a computer's system resources to be exhausted.<br><br>A fork bomb works by repeatedly creating new processes (forks) until the system runs out of resources and becomes unresponsive. The : command is used as a placeholder for the command to be run in the background, and the & operator is used to run the command in the background. The \| operator pipes the output of one command to another command, but in this case, both commands are :, which does nothing. This causes the shell to run an infinite loop, continually creating new processes and consuming more and more system resources.<br><br>It is important to be careful when executing unfamiliar code, especially in a shell, as it could potentially cause harm to the system. |
|---|---|

Task 16 asks ChatGPT to explain the simple robotic navigation code redirecting the robot when it hits a barrier.

**Task 16:** *Revisit the Roomba Vacuum Guidance System (2002), how technology was made to move robotically in the world in complex ways using simple recovery rules*

| **Prompt and code:** please explain this code:<br>  (define-behavior (bounce | **Response:** It is possible that this code could be used to help a robot navigate through its environment. |
|---|---|
| `    :start-when (or (bump?)`<br>`            bounce-trigger? )`<br>`    :abort-when (bump-edge?)`<br>`    :onetime? t`<br>`    )`<br>`)`<br><br>could the logic assist a robot to navigate | The bump? and bump-edge? conditions might be used to detect when the robot has collided with an obstacle or has reached the edge of a surface. The bounce-trigger? condition could be used to trigger the behavior in response to some other event, such as receiving a signal from a sensor.<br><br>The behavior defined by this code might be used to cause the robot to change direction or move away from an obstacle after a collision, or to prevent the robot from falling off the edge of a surface.<br><br>Without more information about the context in which this code is used and the specific capabilities of the robot, it is difficult to provide a more detailed analysis of how it might assist in navigation. |

Task 17 asks ChatGPT to explain the priority algorithm underpinning wireless cell networks.

**Task 17:** *Revisit the Fair Wireless Scheduling (2003), how cellphone networks deconflict interference*

| | |
|---|---|
| **Prompt and code:** How would this code be used to allocate wireless spectrum?<br><br>$[\sim, b\_user] = max(drc(i, :)./avg\_thruput(i, :));$<br>$avg\_thruput(i+1, :) = (i/(i+1))*avg\_thruput(i, :);$<br>$avg\_thruput(i+1, b\_user) = (i/(i+1))*avg\_thruput(i, b\_user)+drc(i, b\_user)/(i+1);$ | **Response:** It looks like this code could be used to allocate wireless spectrum to different users in a dynamic spectrum allocation system. The code appears to be using an algorithm based on the "Exponential-Average Throughput" method, which is a heuristic approach for allocating spectrum to users based on their past throughput (data rate). |

Task 18 asks ChatGPT to explain the bitcoin validation algorithm that rates the (low) probability of blockchain attacks.

**Task 18:** *Revisit Bitcoin (2008), how cryptocurrency reports the small likelihood of an attacker taking over the Bitcoin blockchain*

| Prompt and code: Comment the code: | Glossator: what does the code do in a single sentence? |
|---|---|
| 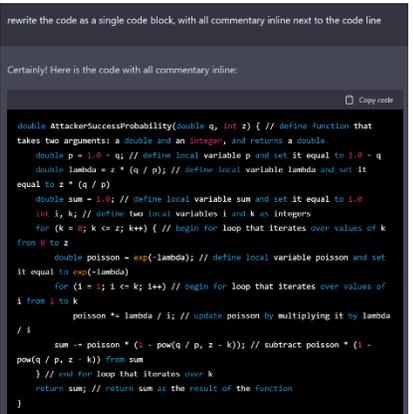 | **Response:** The code defines a function called AttackerSuccessProbability that takes two arguments: a probability q and an integer z, and returns a probability representing the probability of success of an attacker based on the values of q and z.<br>**Glossator:** how would the code assist in the blockchain?<br>**Response:** One possible way in which the code could be used in the context of the blockchain is to calculate the probability of success of an attacker trying to compromise the security of the blockchain network. For example, the q and z arguments might represent the probability of the attacker finding a valid solution to a cryptographic puzzle and the number of attempts the attacker has made, respectively. The return value of the function could then be used to assess the likelihood that the attacker will be able to successfully compromise the network. |

Task 19 asks ChatGPT to explain the Heartbleed security vulnerability, one that had widespread effects but also evaded human detection for a long period.

**Task 19:** *Revisit Heartbleed (2014), how one security vulnerability crippled credentials, encryption keys and 66% of all websites after decades of use*

| **Prompt and code:** Explain what this code might do please:<br><br>```<br>buffer = OPENSSL_malloc(1 + 2 + payload + padding);<br>bp = buffer;<br><br>/* Enter response type, length and copy payload */<br>*bp++ = TLS1_HB_RESPONSE;<br>s2n(payload, bp);<br>memcpy(bp, pl, payload);<br>bp += payload;<br>/* Random padding */<br>RAND_pseudo_bytes(bp, padding);<br>``` | **Response:** This code appears to be part of a program that is using the OpenSSL library to implement the Heartbeat extension of the Transport Layer Security (TLS) protocol. The Heartbeat extension allows a client to send a message to a server, which the server then echoes back to the client, allowing the client to verify that the connection is still active and that the server is responding.<br>Overall, this code appears to be constructing a TLS Heartbeat message and storing it in a block of memory. The message consists of a fixed type field, a variable-length payload field, and a variable-length padding field filled with pseudo-random bytes. The message is stored in a contiguous block of memory, with the various fields being written to the block in sequence.<br>The Heartbeat extension of the TLS protocol, which is implemented using the code you provided, was found to have a vulnerability that allowed an attacker to send a carefully crafted message to a server and potentially reveal sensitive information from the server's memory. |
|---|---|

## 3.2 CODE CORRECTION AND BUG DETECTION DEMONSTRATION

We investigated a second category of code understanding beyond what a traditional Google search might uncover on programming forums. We asked GPT to do obscure code corrections and edge cases such as bug detection. These tasks make up the optimization of SQL to address the problem of missing data or novel queries that the ChatGPT would not encounter in training data.

**Task 20:** *Probe a challenging SQL*

| **Prompt and code:** Using the COALESCE function, add a FullName column using the following query:<br>```sql<br>SELECT<br>     [Title]<br>    ,[FirstName]<br>    ,[MiddleName]<br>    ,[LastName]<br>  FROM [AdventureWorks].[Person].[Person]<br>``` | **Response:** To add a FullName column using the COALESCE function, you can use the following query:<br>```sql<br>SELECT<br>     [Title]<br>    ,[FirstName]<br>    ,[MiddleName]<br>    ,[LastName]<br>    ,COALESCE([Title] + ' ', '') +<br>COALESCE([FirstName] + ' ', '') +<br>COALESCE([MiddleName] + ' ', '') +<br>COALESCE([LastName], '') AS FullName<br>  FROM [AdventureWorks].[Person].[Person]<br>```<br><br>The COALESCE function returns the first non-NULL value in a list of arguments. In this case, it is used to concatenate the values in the Title, FirstName, MiddleName, and LastName columns, with a space character between each column. If any of the columns are NULL, they will be replaced with an empty string, so that the FullName column does not contain any NULL values. The resulting FullName column will contain the concatenated values of the Title, FirstName, MiddleName, and LastName columns, with a space character between each column. |
|---|---|

| **Prompt and code:** Fix this query:<br>```sql<br>SELECT<br>      [Title]<br>      ,[FirstName]<br>      ,[MiddleName]<br>      ,[LastName]<br>      ,COALESCE([Title] + ' ', '') &<br>COALESCE([FirstName] + ' ', '') &<br>COALESCE([MiddleName] + ' ', '') &<br>COALESCE([LastName], '') AS FullName<br>  FROM<br>[AdventureWorks].[Person].[Person]<br>``` | **Response:** It looks like there is a problem with the query because the & operator is being used to concatenate the strings instead of the + operator. The correct query should be:<br><br>```sql<br>SELECT<br>      [Title]<br>      ,[FirstName]<br>      ,[MiddleName]<br>      ,[LastName]<br>      ,COALESCE([Title] + ' ', '') +<br>COALESCE([FirstName] + ' ', '') +<br>COALESCE([MiddleName] + ' ', '') +<br>COALESCE([LastName], '') AS FullName<br>  FROM [AdventureWorks].[Person].[Person]<br>```<br><br>This query will use the + operator to concatenate the values in the Title, FirstName, MiddleName, and LastName columns, with a space character between each column. If any of the columns are NULL, they will be replaced with an empty string, so that the FullName column does not contain any NULL values.<br>The resulting FullName column will contain the concatenated values of the Title, FirstName, MiddleName, and LastName columns, with a space character between each column. |
| **Prompt and code:** Update this query to handle null values for the FullName field.<br><br>```sql<br>SELECT<br>      [Title]<br>      ,[FirstName]<br>      ,[MiddleName]<br>      ,[LastName]<br>      ,[Title] + ' ' + [FirstName] +<br>' ' + [LastName] AS FullName<br>  FROM<br>[AdventureWorks].[Person].[Person]<br>``` | **Response:** To handle NULL values in the FullName field, you can use the COALESCE function as follows:<br><br>```sql<br>SELECT<br>      [Title]<br>      ,[FirstName]<br>      ,[MiddleName]<br>      ,[LastName]<br>      ,COALESCE([Title] + ' ', '') +<br>COALESCE([FirstName] + ' ', '') +<br>COALESCE([LastName], '') AS FullName<br>  FROM [AdventureWorks].[Person].[Person]<br>```<br><br>The COALESCE function returns the first non-NULL value in a list of arguments. In this case, it is used to concatenate the values in the Title, FirstName, and LastName columns, with a space character between each column. If any of the columns are NULL, they will be replaced with an empty string, so that the FullName column does not contain any NULL values.<br>The resulting FullName column will contain the concatenated values of the Title, FirstName, and LastName columns, with a space character between each column. If any of these columns are NULL, they will be replaced with an empty string in the FullName column. |
| **Prompt and code:** Extract the Access Query from this text:<br>```<br>'Status disagrees<br> strSQLStatement = "UPDATE PartData<br>INNER JOIN PartStatus " & _<br>    "ON PartData.PartNumber =<br>PartStatus.PartNumber " & _<br>       "AND PartData.Manufacturer =<br>PartStatus.Manufacturer " & _<br>    "SET PartData.Status =<br>'Incorrect', " & _<br>       "PartData.Comments = 'Tools<br>Disagree'" & _<br>``` | **Response:** Here is the Access query in plain text:<br><br>```sql<br>UPDATE PartData INNER JOIN PartStatus<br>ON PartData.PartNumber =<br>PartStatus.PartNumber<br>AND PartData.Manufacturer =<br>PartStatus.Manufacturer<br>SET PartData.Status = 'Incorrect',<br>PartData.Comments = 'Tools Disagree'<br>WHERE ((PartData.VendorBStatus IS NOT NULL<br>And PartData.VendorBStatus <> '')<br>       And (PartData.VendorAStatus IS NOT<br>NULL And PartData.VendorAStatus <> ''))<br>AND (PartData.Status =<br>PartStatus.FinalStatus)<br>``` |
| `        "WHERE ((PartData.VendorBStatus IS NOT NULL And PartData.VendorBStatus <> '') " & _`<br>`            "And (PartData.VendorAStatus IS NOT NULL And PartData.VendorAStatus <> '')) " & _`<br>`        "AND (PartData.Status = PartStatus.FinalStatus) "` | |
| **Prompt and code:** Convert that to work in SQL 2016 | **Response:** Here is the query converted to work in SQL Server 2016:<br><br>`UPDATE pd`<br>`SET pd.Status = 'Incorrect',`<br>`    pd.Comments = 'Tools Disagree'`<br>`FROM PartData pd`<br>`INNER JOIN PartStatus ps`<br>`    ON pd.PartNumber = ps.PartNumber`<br>`    AND pd.Manufacturer = ps.Manufacturer`<br>`WHERE (pd.VendorBStatus IS NOT NULL AND pd.VendorBStatus <> '')`<br>`    AND (pd.VendorAStatus IS NOT NULL AND pd.VendorAStatus <> '')`<br>`    AND (pd.Status = ps.FinalStatus)` |
| **Prompt and code:** Optimize the WHERE clause | **Response:**<br>Here are a few suggestions for optimizing the WHERE clause in the query:<br>Consider using the COALESCE() function to check for NULL values in the VendorAStatus and VendorBStatus columns. This can avoid having to use multiple AND clauses to check for NULL and empty values.<br>`WHERE COALESCE(pd.VendorAStatus, '') <> ''`<br>`    AND COALESCE(pd.VendorBStatus, '') <> ''`<br>Use the EXISTS operator to check for the existence of rows in the PartStatus table that match the PartNumber and Manufacturer columns in the PartData table. This can improve the performance of the query, especially if the PartStatus table is large.<br>`WHERE EXISTS (`<br>`    SELECT 1`<br>`    FROM PartStatus ps`<br>`    WHERE pd.PartNumber = ps.PartNumber`<br>`        AND pd.Manufacturer = ps.Manufacturer`<br>`)`<br>Consider using table indexes to improve the performance of the JOIN and WHERE clauses. This can make the query run faster by allowing the database server to quickly locate the rows it needs to update.<br><br>`CREATE INDEX idx_PartNumber_Manufacturer ON PartStatus (PartNumber, Manufacturer);`<br><br>Here is the revised query using these suggestions:<br>`UPDATE pd`<br>`SET pd.Status = 'Incorrect',`<br>`    pd.Comments = 'Tools Disagree'`<br>`FROM PartData pd`<br>`INNER JOIN PartStatus ps`<br>`    ON pd.PartNumber = ps.PartNumber`<br>`    AND pd.Manufacturer = ps.Manufacturer` |

```
                                           WHERE COALESCE(pd.VendorAStatus, '') <> ''
                                                 AND COALESCE(pd.VendorBStatus, '') <>
                                           ''
                                                 AND EXISTS (
                                                     SELECT 1
                                                     FROM PartStatus ps
                                                     WHERE pd.PartNumber =
                                           ps.PartNumber
                                                         AND pd.Manufacturer =
                                           ps.Manufacturer
                                                     )
                                                 AND (pd.Status <> ps.FinalStatus)
```

## 4. DISCUSSION AND CONCLUSIONS

The research has applied AI-driven code assistants to explore half of the surveyed computer code that changed the course of modern technology, including email, internet browsing, robotics, and malicious software. The original contribution of this research analyzes half of the most significant code advances in the last half-century and, in some cases, offers considerable improvements in clarity or performance. In no instances examined was ChatGPT unable or unwilling to shed light on obfuscated code or software lacking explanatory commentary. Some benchmark tasks, such as the Apollo Lunar Lander memory management software, would strain the best modern software analysts. We generated additional sample problems based on bug corrections and code optimizations requiring much deeper reasoning than a traditional Google search might provide. Future work will emphasize adding automated documentation and code commentary along with translating selective large code bases into more modern versions that make contact with multiple new application programming interfaces (API) and chained multi-tasks. In software engineering, the advent of human-level software expertise promises new approaches to refactoring legacy code or simplifying the explanation or overall functionality of high-value repositories. Akin to a new scientific instrument, the AI-driven code assistant seems fruitful for further practical exploration, particularly in its most recent incarnation as a chat interface capable of long-term memory, context, and polyglot understanding of software principles learned by examples without explicit rules or human guidance.

## ACKNOWLEDGEMENTS

The authors thank the PeopleTec Technical Fellows program for encouragement and project assistance. The authors thank the researchers at OpenAI for developing large language models and allowing public access to ChatGPT.

## Authors

**David Noever** has research experience with NASA and the Department of Defense in machine learning and data mining. He received his BS from Princeton University and his Ph.D. from Oxford University, as a Rhodes Scholar, in theoretical physics.

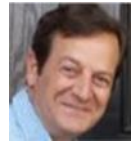

**Kevin Williams** has research experience in software engineering and development. He received his Bachelor's in Computer Science from Troy University and his Master's in Software Engineering from Pennsylvania State University.

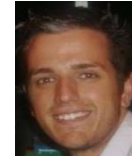

> **Technical Note: Some appendix text generated from Large Language Model (LLM) for illustration purposes.**
>
> *The author generated this text in part with GPT-3, OpenAI's large-scale language-generation model. Upon generating draft language, the authors reviewed, edited, and revised the language to their own liking and take ultimate responsibility for the content of this publication.*
>
> OpenAI policy statement (2022)

## APPENDIX A: CHATGPT SUMMARIZATION OF 1961 SPACEWARS! CODE

```
Prompt: What does this code do?
/ this routine handles a non-colliding ship invisibly
/ in hyperspace
hp1,  dap hp2
  count i ma1, hp2
  law hp3    / next step
  dac i ml1
  law 7
  dac i mb1
  random
  scr 9s
  sir 9s
  xct hr1
  add i mx1
  dac i mx1
  swap
  add i my1
  dac i my1
  random
  scr 9s
  sir 9s
  xct hr2
  dac i mdy
  dio i mdx
  setup .hpt,3
  lac ran
  dac i mth
```

```
hp4,   lac i mth
       sma
       sub (311040
       spa
       add (311040
       dac i mth
       count .hpt,hp4
       xct hd2
       dac i ma1
hp2,   jmp .
```
ChatGPT: It looks like this is a code written in the Assembly programming language. It appears to be a routine that handles a non-colliding ship while in hyperspace. It appears to be updating the position of the ship, possibly based on randomness, and checking if certain conditions are met. Without more context, it is difficult to say exactly what the code is doing. Can you provide more information about the code or its intended purpose?

Prompt: If plays spacewars in which two players control spaceships, the needle and the wedge, and engage in a one-on-one space battle while avoiding a gravity well of a star at the center of the screen.

ChatGPT: Based on this additional context, it looks like the code you provided is handling the movement of one of the spaceships in the game, possibly the "needle" spaceship. It appears to be using some randomness in its movement, and there are references to certain conditions (such as "ml1" and "mb1") that may be relevant to the game's mechanics. It is also possible that the code is interacting with other parts of the game, such as handling the display of the spaceship or processing input from the player controlling the spaceship. Without more information about the specific game mechanics and the purpose of the different variables and instructions in the code, it is difficult to say exactly what the code is doing.

Prompt: can you translate it to a python spacewars game doing approximately the same task

ChatGPT: Sure, here is a rough translation of the code you provided into Python that accomplishes roughly the same task:

```python
import random
def handle_noncolliding_ship_in_hyperspace(ma1, hp2, ml1, mb1, mx1, my1, hr1,
hr2, mdy, mdx, hd2):
  for i in range(hp2):
    if random.random() < 0.5:
      mx1 += ml1
    else:
      mb1 += 7
    mx1 += hr1
    random.seed(9)
    mx1, my1 = my1, mx1
    random.seed(9)
    mdy, mdx = hd2()
  setup_hpt(3)
  mth = random.random()
  while True:
    mth += (random.random() - 0.5) * 2
    if mth < -311040 or mth > 311040:
      mth -= (random.random() - 0.5) * 2
    hpt_count += 1
    if hpt_count >= hp4:
      break
  ma1 = hd2()
```

This translation is a rough approximation of the original code, and it may not be fully functional without further context about the specific game mechanics and the purpose of the different variables and instructions in the original code.

Prompt: can you make it output some ascii representation of the needle and wedge?

**ChatGPT:** Sure, here is an example of how you could output an ASCII representation of the "needle" spaceship using Python:

```
def print_needle():
  print("  /\\")
  print(" /  \\")
  print("/____\\")
  print("|    |")
  print("|    |")
  print(" \\  /")
  print("  \\/")
print_needle()
```

This would output the following ASCII art:

```
  /\
 /  \
/____\
|    |
|    |
 \  /
  \/
```

You could use a similar approach to output an ASCII representation of the "wedge" spaceship, or you could use a different ASCII art design if you prefer.

```
def print_wedge():
  print("    /\\")
  print("   /  \\")
  print("  /____\\")
  print(" / /  \\ \\")
  print("/ /____\\ \\")
  print("   |    |")
  print("   |    |")
  print("    \\  /")
  print("     \\/")
print_wedge()
```

This would output the following ASCII art:

```
    /\
   /  \
  /____\
 / /  \ \
/ /____\ \
   |    |
   |    |
    \  /
     \/
```

## APPENDIX B: CHATGPT SUMMARIZATION OF EMAIL AGENT CODE

**Prompt:** Explain what this code does
WHENEVER A(1).E.FENCE.OR.A(2).E.FENCE.OR.A(3).E.FENCE
PRFULL.($'R'1INSTRUCTIONS:$)
PRFULL.($ '4MAIL NAME1 NAME2 PROB1 PROG1 PROB2 PROG2 ...$)
PRFULL.($ WHERE '=NAME1 NAME2'= IS THE FILE TO BE MAILED,$)

PRFULL.($ AND '=PROBN PROGN'= ARE DIRECTORIES TO WHICH '8$,
1 $IT IS TO BE SENT.'B$)
CHNCOM.(0)
END OF CONDITIONAL

**ChatGPT:** It looks like this is a code written in a variant of the COBOL programming language. The code appears to be defining a conditional statement that will be executed whenever the condition "A(1).E.FENCE.OR.A(2).E.FENCE.OR.A(3).E.FENCE" is true. The condition appears to be checking if certain values ("A(1)", "A(2)", "A(3)") are equal to the value "FENCE". If the condition is true, the code will execute the following three statements:

"PRFULL.($'R'1INSTRUCTIONS:$)" - This statement appears to be using the "PRFULL" function to print a string containing the text "INSTRUCTIONS:".
"PRFULL.($ '4MAIL NAME1 NAME2 PROB1 PROG1 PROB2 PROG2 ...$)" - This statement appears to be using the "PRFULL" function to print a string containing multiple items, possibly separated by spaces or other delimiters.
"PRFULL.($ WHERE '=NAME1 NAME2'= IS THE FILE TO BE MAILED,$) PRFULL.($ AND '=PROBN PROGN'= ARE DIRECTORIES TO WHICH '8$, 1 $IT IS TO BE SENT.'B$)" - This statement appears to be using the "PRFULL" function to print a string containing multiple lines of text that provide further instructions or details about the operation being performed.
The code also includes a statement "CHNCOM.(0)" which appears to be calling a function "CHNCOM" with the argument "0". It is not clear what this function does or what the purpose of this statement is. Finally, the code includes the statement "END OF CONDITIONAL" which indicates the end of the conditional block that was defined at the beginning of the code.

**Prompt:** translate the code to python to accomplish a similar thing

ChatGPT: Here is a rough translation of the code you provided into Python that accomplishes roughly the same thing:

if A[1] == "FENCE" or A[2] == "FENCE" or A[3] == "FENCE":
  print("INSTRUCTIONS:")
  print("MAIL NAME1 NAME2 PROB1 PROG1 PROB2 PROG2 ...")
  print("WHERE '=NAME1 NAME2'= IS THE FILE TO BE MAILED,")
  print("AND '=PROBN PROGN'= ARE DIRECTORIES TO WHICH '8$, 1 $IT IS TO BE SENT.'B$")
  CHNCOM(0)

This translation is a rough approximation of the original code, and it may not be fully functional without further context about the specific purpose and behavior of the variables, functions, and statements in the original code. In particular, the function "CHNCOM" and its purpose are not clear from the original code, so it is not included in the translation.

**Prompt:** its an electronic mail system

**ChatGPT**: Based on this additional context, it looks like the code you provided is part of an electronic mail system. The code defines a conditional statement that is executed when certain conditions are met (i.e. when the values of "A[1]", "A[2]", or "A[3]" are equal to "FENCE"). When the condition is true, the code appears to be printing some instructions and possibly some data related to mailing a file to one or more directories. The function "CHNCOM" is not mentioned in the original code, so it is not clear what it does or how it is used in the mail system. Without more information about the specific implementation and behavior of the mail system, it is difficult to say exactly what the code is doing or how it fits into the overall system.